\documentclass[runningheads]{llncs}
\usepackage{graphicx}
\usepackage{amsmath}

\begin{document}
%===========================================================
\title{ColorNet: Investigating the importance of color spaces for image classification\thanks{Supported by NSFC project Grant No. U1833101, Shenzhen Science and Technologies project under Grant No. JCYJ20160428182137473 and the Joint Research Center of Tencent and Tsinghua.}} % Replace your paper's title here
\titlerunning{Colornet} % Replace an abstracted version of your paper's title here

%===========================================================

\author{Shreyank N Gowda\inst{1} \and
Chun Yuan\inst{2}\
}
%
%Please include author names in full in the paper, 
%If any authors have names that can be parsed into FirstName LastName in multiple ways, please include the correct parsing, in a comment to the volume editors:
%\index{Lastnames, Firstnames}

\authorrunning{Shreyank N Gowda et al.} % A shorter version of authors' name

%===========================================================

\institute{Computer Science Department, Tsinghua University, Beijing 10084, China \email{sny17@mails.tsinghua.edu.cn} \and
Graduate School at Shenzhen, Tsinghua University, Shenzhen 518055, China
\email{yuanc@sz.tsinghua.edu.cn
}\\
}

\maketitle

%===========================================================
\begin{abstract}
Image classification is a fundamental application in computer vision. Recently, deeper networks and highly connected networks have shown state of the art performance for image classification tasks. Most datasets these days consist of a finite number of color images. These color images are taken as input in the form of RGB images and classification is done without modifying them. We explore the importance of color spaces and show that color spaces (essentially transformations of original RGB images) can significantly affect classification accuracy. Further, we show that certain classes of images are better represented in particular color spaces and for a dataset with a highly varying number of classes such as CIFAR and Imagenet, using a model that considers multiple color spaces within the same model gives excellent levels of accuracy. Also, we show that such a model, where the input is preprocessed into multiple color spaces simultaneously, needs far fewer parameters to obtain high accuracy for classification. For example, our model with 1.75M parameters significantly outperforms DenseNet 100-12 that has 12M parameters and gives results comparable to Densenet-BC-190-40 that has 25.6M parameters for classification of four competitive image classification datasets namely: CIFAR-10, CIFAR-100, SVHN and Imagenet. Our model essentially takes an RGB image as input, simultaneously converts the image into 7 different color spaces and uses these as inputs to individual densenets. We use small and wide densenets to reduce computation overhead and number of hyperparameters required. We obtain significant improvement on current state of the art results on these datasets as well.

\keywords{Color spaces  \and Densenet \and Fusion.}
\end{abstract}
%===========================================================
\section{Introduction}

Image classification is one of the most fundamental applications in the field of computer vision. Most of the datasets used for image classification tend to consist of color images. These color images are represented in RGB format. To a computer, these images are just numbers and do not contain any inherent meaning. Most recent models developed for classification do not perform a color space transformation to the image and instead use the RGB image directly for classification.

In this paper, we propose the use of different color spaces. The main color spaces will be discussed briefly in this section.

A color space is essentially an organization of colors. Along with physical device profiling, color spaces help us to reproduce analog and digital representations of color. Color spaces can also be thought of as an abstract mathematical model that helps us to describe colors as numbers.

RGB color space, often the most popular color space, is a system dependent color space. Commonly, it is represented using a 24-bit implementation where each channel R, G and B are given 8 bits each. This results in each channel having a range of values from 0 to 255. This color space models on the basis that all colors can be represented using different shades of red, green and blue.

Images from popular datasets such as CIFAR [2] have images present in the sRGB format. The first step is in converting the sRGB
images to RGB by linearizing it by a power-law of 2.2. 

Some of the other popular color spaces we shall discuss briefly are HSV, LAB, YUV, YCbCr, YPbPr, YIQ, XYZ, HED, LCH and CMYK.

HSV stands for hue, saturation and value. HSV was developed taking into consideration how humans view color. It describes color (hue) in terms of the saturation (shade) and value (brightness). H has a range from 0 to 360, S and V have range 0 to 255. The transformation from RGB to HSV can be seen in (1)-(6). R, G and B are the values of the red channel, green channel and blue channel respectively. H obtained represents the hue channel. Similarly, S represents the saturation channel and V the value channel.
\begin{align}
  R' = R/255, G' = G/255, B' = B/255 \; \\
  Cmax = max(R', G', B'), Cmin = min(R', G', B') \; \\
   \Delta = Cmax - Cmin \; \\
   H=\left\{\begin{matrix}
0^{\circ} & ,\Delta = 0\\ 
60^{\circ}\left ( \frac{G'-B'}{\Delta} mod6\right ) & ,Cmax=R'\\ 
60^{\circ}\left ( \frac{B'-R'}{\Delta} +2\right ) & ,Cmax=G'\\ 
60^{\circ}\left ( \frac{R'-G'}{\Delta} + 4\right ) & ,Cmax=B'
\end{matrix}\right. \; \\
S = \left\{\begin{matrix}
0 & ,Cmax=0\\ 
\frac{\Delta}{Cmax} & ,Cmax \neq 0
\end{matrix}\right. \; \\
V = Cmax
\end{align}

To define quantitative links between distributions of wavelengths in the EM visible spectrum (Electromagnetic) along with the physiological perceived colors in human eye sight, the CIE (Commission internationale de l’éclairage) 1931 color spaces were introduced. The mathematical relationships between these color spaces form fundamental tools to deal with color inks, color management, illuminated displays, cameras and printers.

CIE XYZ (from now on referred to as XYZ) was formed on the mathematical limit of human vision as far as color is concerned. X, Y and Z are channels extrapolated from the R, G and B channels to prevent the occurrence of negative values. Y represents luminance, Z represents a channel close to blue channel and X represents a mix of cone response curves chosen to be orthogonal to luminance and non-negative. XYZ image can be derived from RGB using (7).
\begin{align}
 \begin{pmatrix}
X\\ 
Y\\ 
Z
\end{pmatrix}
=
\begin{bmatrix}
0.489989 & 0.310008  &  0.2 \\ 
0.176962 & 0.81240 & 0.010\\ 
0 & 0.01 & 0.99
\end{bmatrix}
\begin{pmatrix}
R\\ 
G\\ 
B
\end{pmatrix}
\end{align}

LAB is another very popular color space. This color space is often used as an interchange format when dealing with different devices. This is done because it is device independent. Here, L stands for lightness, 'a' stands for color component green-red and 'b' for blue-yellow. An image in RGB can be transformed to LAB by first converting the RGB image to XYZ image.

YIQ, YUV, YPbPr, YCbCr are all color spaces that are used for television transmission. Hence, they are often called transmission primaries. YUV and YIQ are analog spaces for PAL and NTSC systems. YCbCr is used for encoding of digital color information used in video and still-image transmission and compression techniques such as JPEG and MPEG.

YUV, YCbCr, YPbPr, and YIq belong to opponent spaces. They have one channel for luminance and 2 channels for chrominace, represented in an opponency way (usually red versus green, and blue versus yellow).

The RGB color space is an additive color model, where to obtain the final image we add the individual channel values. A subtractive color model exists, called CMYK where C stands for Cyan, M stands for magenta, Y stands for yellow and K stands for key (black). The CMYK model works by masking colors on a light background. RGB to CMYK conversion can be seen in (8)-(12).

\begin{align}
  R' = R/255, G' = G/255, B' = B/255 \; \\
  K = 1-max(R', G', B') \; \\
   C = \frac{(1-R'-K)}{(1-K)}\; \\ M = \frac{(1-G'-K)}{(1-K)} \; \\ Y = \frac{(1-B'-K)}{(1-K)}
\end{align}

LCH is another color space. It is similar to LAB. It is in the form of a sphere that has three axes: L, c and h. L stands for lightness, c stands for chroma (saturation) and h stands for hue. It is a device-independent color space and is used for retouching images in a color managed workflow that utilizes high-end editing applications. To convert from RGB to LCH, we first convert from RGB to LAB and then from LAB to LCH.

There are other color spaces that have not been explored. Only the most popular ones have been referenced. The idea behind the paper is that an image is nothing but numbers to a computer, hence, transformations of these images should be viewed as a completely new image to a computer. Essentially transforming an image into different color spaces should yield us new images in the view of the computer.

We exploit our idea by using small networks to classify images in different color spaces and combine the final layer of each network to obtain an accuracy that takes into account all the color spaces involved. To obtain a high accuracy, we need each output to be less correlated to each other. This is also something we will explain in detail in the proposed approach section.

We have proposed the following novel contributions, that have not been performed before to the best of our knowledge.

1) We show that certain classes of images are better represented in certain color spaces.

2) We show that combining the outputs for each color space will give us a much higher accuracy in comparison to individually using each color space.

3) We show that a relatively small model can obtain a similar level of accuracy to recent state of the art approaches (See experimental analysis section, our model with 1.75M parameters can provide a similar accuracy to that of a densenet model with 25.6M parameters for CIFAR datasets)

4) We also obtain new state of the art results on CIFAR-10, CIFAR-100, SVHN and imagenet, to the best of our knowledge.

\section{Related Works}

Image classification has been a fundamental task in the field of computer vision. This task has gained huge importance in recent times with the development of datasets such as Imagenet [10, CIFAR [2], SVHN[3], MNIST [4], CALTECH-101 [5], CALTECH-256 [6] among others.

Deep convolutional neural networks[7-9] have been developed which have drastically affected the accuracy of the image classification algorithms. These have in turn resulted in breakthroughs in many image classification tasks [10-11].

Recent research[12-13] has shown that going deeper will result in higher accuracy. In fact, current state-of-the-art approaches on the challenging Imagenet [1] dataset has been obtained by very deep networks [12-15]. Many complex computer vision tasks have also been shown to obtain great results on using deeper networks [16-20].

Clearly, depth has been giving us good results. But learning better networks is not as simple as adding more layers. A big deterrent for this is the vanishing gradient problem which hampers convergence of layers [21].

A solution to the vanishing gradient problem has been normalized initialization [22,23]. Another solution has been the introduction of intermediate normalization layers[15]. These enabled very deep networks to start converging for SGD (stochastic gradient descent) with back propagation [24].

Networks with an extremely high number of layers were giving high accuracy results on Imagenet as mentioned before. For example, [25] had 19 layers, [26-27] surpassed 100 layers. Eventually, it was seen in [28] that not going deeper, but going wider provided a higher accuracy.

The problem with deeper networks as mentioned above is the vanishing gradient problem. To solve this we could bypass signal between layers using identity connections as seen in the popular networks Resnets[26] and Highway networks[27].

Repeatedly combining multiple parallel layer sequences with a varying number of convolutional blocks was done in order to get a large nominal depth in FractalNets [29]. This was done along with maintaining multiple short paths in the network. The main similarity between [26-29] was that they all created short paths from early layers to the later layers.

Densenet [30] proposed connecting all layers to ensure information from each layer is passed on to every other layer. They also showed state-of-the-art results on popular image classification tasks.

All these recent approaches gave excellent results, however, the number of parameters has been very high. We look at the possibility of reducing the number of parameters needed for performing the same task, whilst, ensuring the accuracy of classification remains high.

Also, all these recent approaches used images from the dataset directly as it is for the task of classification. We propose transformations of these images using color space conversions as the medium to do so.

Performing image classification tasks by preprocessing input with color conversion has been explored before. In [31] for instance, YCbCr was the color space used for skin detection. Color pixel classification was done in [32] using a hybrid color space. Soccer image analysis was done using a hybrid color space in [33]. 

To see if color space conversion actually makes a difference, an analysis was done on skin detection [34]. They found that RGB color space was the model that gave the best results. Skin pixel classification was studied using a Bayesian model in [35] with different color spaces. In this case, the authors found LAB color space gave best results for accuracy.

Based on the works in [34-35] we can say that the authors found conflicting results. But importantly we take from these works the fact that using different color spaces gave authors different results, which means we could experiment on the same. We exploit this idea for our approach. The main modification we do, which will be explained in the proposed approach section, is that we combine the color spaces model to obtain a higher accuracy. This is due to the fact, as we shall see in the next section, that the color spaces are not completely correlated.
\section{Proposed Approach}

The idea was thought of while trying the effects of color space conversion on the CIFAR-10 dataset [2]. We started with a simple convolutional network. The network consisted of two convolutional layers followed by a max pooling layer. This was followed by a dropout layer, 2 more convolutional layers, one more max pooling, one more dropout layer and finished with a dense layer.

We started the classification on CIFAR-10 using the input data as it is i.e in RGB format. We obtained an accuracy of 78.89 percent and time needed was 26 seconds. Next, we performed the classification by introducing color space conversion. We did the same with HSV, LAB, YIQ, YPbPr, YCbCr, YUV, LCH, HED, LCH and XYZ. The reason we chose the selected color spaces was due to the ease to perform the conversion using Scikit library [36]. The results of the classification can be seen in Table 1.

\setlength{\tabcolsep}{4pt}
\begin{table}[h]
\begin{center}
\caption{Comparison of results for different color spaces on CIFAR-10 with simple CNN}
\label{table:simpleCNN}
\begin{tabular}{lll}
\hline\noalign{\smallskip}
Color Space & Accuracy & Time\\
\noalign{\smallskip}
\hline
\noalign{\smallskip}
RGB  & 78.89 & 26 secs\\
HSV  & 78.57 & 26 secs\\
YUV  & 78.89 & 26 secs\\
LAB  & \textbf{80.43} & 26 secs\\
YIQ  & 78.79 & 26 secs\\
XYZ  & 78.72 & 26 secs\\
YPbPr  & 78.78 & 26 secs\\
YCbCr  & 78.81 & 26 secs\\
HED  & 78.98 & 26 secs\\
LCH & 78.82 & 26 secs\\

\hline
\end{tabular}
\end{center}
\end{table}
\setlength{\tabcolsep}{1.4pt}

From the table, we can see that the results obtained were highest for LAB color space, whilst the time of execution remained constant. However, like previous papers [34-35] showed, the accuracy levels are not too distant showing that the color space conversion results in more or less the same results while adding the time needed for the conversion.

We decided to have a closer look at the results by observing the confusion matrix for each case. The confusion matrices are shown in Figure 1 for RGB, HSV, YUV, LAB, HED, LCH, XYZ and YPbPr.

\begin{figure}[h]
\begin{center}
   \includegraphics[width=1.0\linewidth]{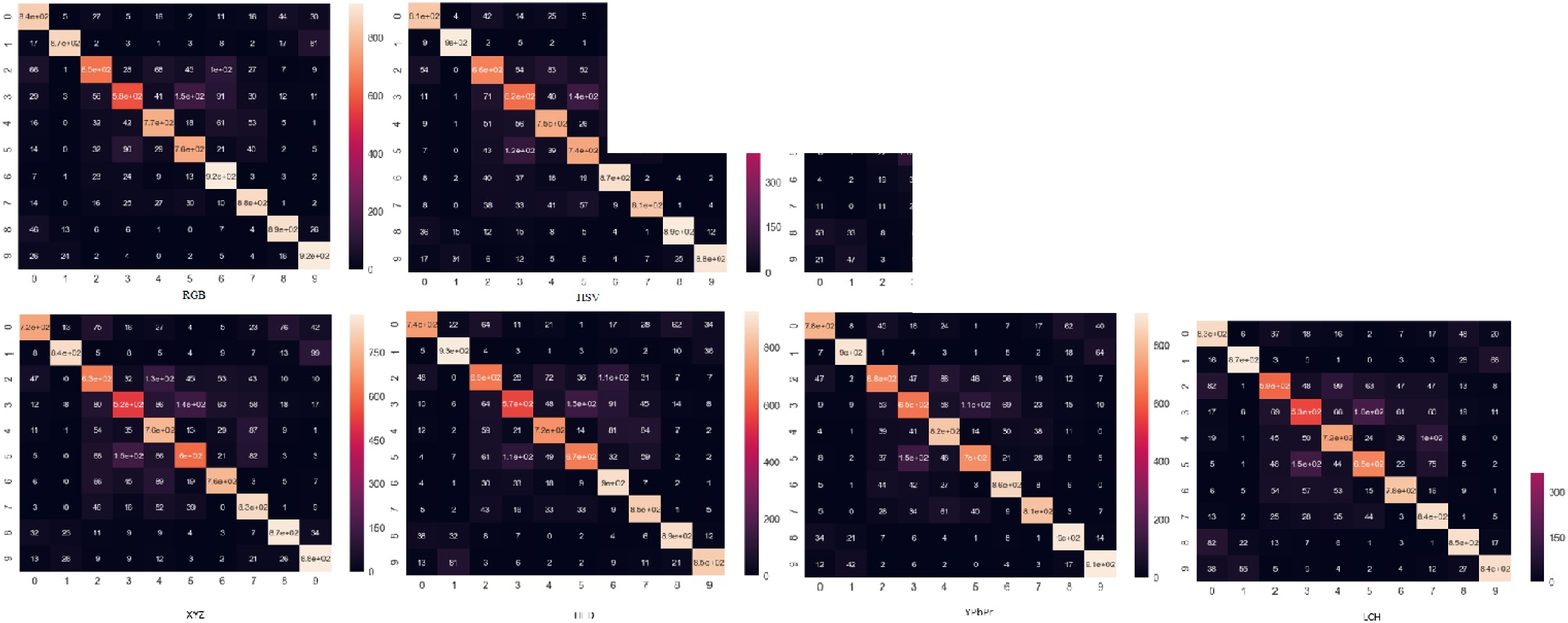}
\end{center}
   \caption{Confusion matrices for various color spaces}
\label{fig:cm}
\end{figure}

On closer inspection, we see that each color space gives different accuracy for different classes. There are some classes that have equal classification rate in some classes, however, the majority differ in some way. For example, class 4 in CIFAR-10 is detected with 82 percent accuracy using YPbPr, however, the same class is detected only with 72 percent in case of HED. Such differences can be noticed in multiple classes.

Based on these findings, we can conclude two important things. Firstly, there isn't 100 percent correlation between color spaces, which infers they can be used in combination to obtain better results. Secondly, certain classes of objects are being better represented by different color spaces.

We use these findings as the basis of our proposed model. We use small and wide Densenets as the base of our combination model. The proposed model is explored in the next section. Another important finding was that some color spaces are causing a loss of accuracy. We show a sample of such losses in Table 2. Here, we individually combined multiple color spaces to obtain higher classification rates using the simple CNN proposed earlier. This table was created only to show the reason we exclude certain color spaces in the next section. For this, we use RGB, HSV, XYZ, HED, LAB and YUV.

We also tried a late fusion approach. This would decrease the number of parameters in the model.

\setlength{\tabcolsep}{4pt}
\begin{table}[h]
\begin{center}
\caption{Comparison of results for combination of different color spaces on CIFAR-10 with simple CNN}
\label{table:simpleCNNCombination}
\begin{tabular}{lll}
\hline\noalign{\smallskip}
Color Space & Accuracy & Number of color spaces used\\
\noalign{\smallskip}
\hline
\noalign{\smallskip}
RGB+HSV	& 81.42 & 2\\
RGB+YUV	& 81.41 & 2\\
HSV+YUV	& 81.97 & 2\\
RGB+LAB	& 81.91 & 2\\
LAB+HSV	& 81.95 & 2\\
YUV+LAB	& 82.05 & 2\\
RGB+HSV+YUV	& 82.33 & 3\\
RGB+HSV+LAB	& 82.49 & 3\\
RGB+LAB+YUV	& 82.62 & 3\\
LAB+HSV+YUV	& 82.66 & 3\\
RGB+HSV+YUV+LAB	& 82.96 & 4\\
RGB+HSV+YUV+LAB+HED(RHYLH)	& 83.61 & 5\\
RGB+HSV+YUV+LAB+HED+XYZ	& 82.81 & 6\\
RHYLH with Conv Layer	& \textbf{84.32} & 5\\
RHYLH with early fusion & 81.63 & 5\\

\hline
\end{tabular}
\end{center}
\end{table}
\setlength{\tabcolsep}{1.4pt}

The combination being talked about above is using the output of the simple CNN with a particular color space and combining the outputs to obtain an average output of the different color spaces. The idea will be better understood in the next subsection. However, the important finding was that we can use the combination of color spaces to obtain a high accuracy of classification.

As can be seen, an early fusion approach did not significantly alter the accuracy in comparison to a model with a single color space itself. Although, the number of parameters significantly reduced in comparison to the model with late fusion, the accuracy was also lower in case of early fusion. Hence, we have chosen a late fusion model.

\subsection{Architecture of model used}

The base of the model is using a Densenet for each color space. Essentially, the model consists of 7 Densenets, one for each color space being used, along with a final dense layer to give weighted predictions to classes.

Densenet [30] proposed the use of feed-forward connection between every layer in a network. They proposed the idea to overcome problems such as the vanishing gradient problem, but also believed that such a network could obtain remarkably high results with fewer parameters in comparison to other recent models. The Densenet model is represented as DenseNet-L-k where L stands for the depth of the model and k stands for the growth factor.

However, the model that obtained highest accuracy in both CIFAR-10 and CIFAR-100 datasets, the DenseNet-190-40 needed 25.6M parameters. We believed that this could be reduced significantly if we could find a way to preprocess the data.

We notice from our observations in the previous section, that combining the outputs from the 7 CNNs increased the accuracy of the model from 78.89 percent to 86.14 percent. Which meant a rise close to 7 percent. This result provoked the thought that, using a combination of smaller densenets which use far fewer parameters than the ones that obtained state of the art results could help us to reach similar levels of accuracy if not obtain better ones. Along with this, we thought of the idea that a wider densenet could possibly obtain better results than a deeper densenet based on [28].

Using these two thoughts, we decided to implement a model that consisted of small densenets for each color space. We started by using a DenseNet-BC-40-12 which used only 0.25M parameters. The BC in the name refers to the use of bottleneck layers after each block to compress the size of the model.

The color spaces that were selected based on the results obtained by individually checking if the accuracy affects the overall model are: RGB, LAB, HSV, YUV, YCbCr, HED and YIQ. The overall architecture of the model is seen in Figure 2.

\begin{figure}[h]
\begin{center}
   \includegraphics[width=1.0\linewidth]{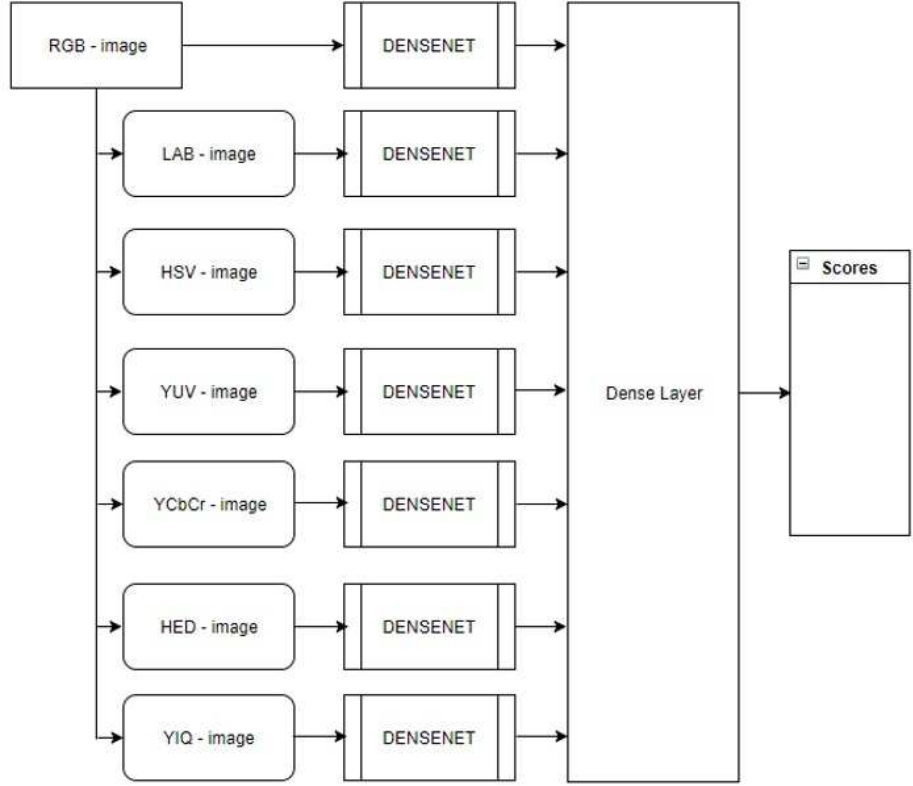}
\end{center}
   \caption{Architecture of proposed model}
\label{fig:cm}
\end{figure}

The input RGB image is simultaneously converted into the 6 other color spaces and all 7 inputs are passed to Densenets. The output scores of each Densenet is then passed to a dense layer which helps to give weighted predictions to each color space. The output of the dense layer is used as the final classification score.

Many questions may arise for such a model, for example, the time needed for multiple color space conversions will cause an overhead. Also, the Densenets as mentioned in earlier sections has many parameters by itself, therefore, a combination of densenets will have, in this case, 7 times the number of parameters.

The first problem is something we do for the benefit of a higher accuracy which will be seen in later sections. The second, however, is solved by using smaller and wider densenets which use far fewer parameters than the models that have state of the art results on popular image classification datasets. The experimental results should help satisfy some of the questions regarding the time-accuracy trade-off.
\section{Experimental Analysis}

\subsection{Datasets}

We perform the experimental evaluation on CIFAR-10, CIFAR-100, SVHN and Imagenet.

The CIFAR-10 dataset consists of 60000 color images of size 32x32 each. There are 10 classes present. The dataset is divided into 50000 training images and 10000 test images. The CIFAR-100 dataset consists of 100 classes. We use data augmentation for these datasets and represent the results with a '+' sign. For example, C10+ refers to Cifar-10 results with augmentation. The augmentation done is horizontal flip and width/height shift.

The SVHN dataset (Street View House Numbers) contains 32x32 size images. There are 73,257
training images in the dataset and 26,032 test images in the dataset. Additionally, there are 531,131 images for extra training.

The Imagenet dataset consists 1.2 million training images and 50000 validation images. There are a total of 1000 classes. We apply single crop or 10 crop with size 224x224 at test time. We follow [27] and [30] in showing the imagenet classification score on the validation set.

\subsection{Training}

Since we use densenets as the base for building our model, we follow the training procedure followed in [30]. Each individual densenet is trained using stochastic gradient descent (SGD). We use a batch size of 32 for CIFAR datasets and run the model for 300 epochs. The learning rate is initially 0.1 and is reduced to 0.01 after 75 epochs and 0.001 after 150 epochs. In case of SVHN we run the model for 100 epochs and the learning rate is initially 0.1 and is reduced to 0.01 after 25 epochs and 0.001 after 50 epochs.

As with the original densenet implementation, we use a weight decay of 0.0001 and apply Nesterov momentum [37] of 0.9 without dampening. When we do not use data augmentation we add a dropout of 0.2.

There are 2 parameters that can affect the accuracy and number of parameters of our model. These are the growth factor and depth of network as was the case with original densenet [30]. With this regard, here onwards, colornet-L-k refers to a colornet that has densenets-L-k as subparts of the colornet model. L here refers to the depth of the network and k the growth factor. We use densenets with bottleneck as the densenet model part of our model.

\subsection{Classification results on CIFAR-10}

Table 3 refers to the classification accuracies obtained on CIFAR-10 dataset for recent state of the art approaches. We compare the results obtained from our model with these approaches. In the table C10 refers to the accuracy of a particular model on CIFAR-10 and C10+ refers to the accuracy of the same model with data augmentation.

We compare our model with Network in Network [38], the All-CNN [39], Highway network [26], fractalnet [29], Resnet [27], Wide-Resnet [28] and Densenet [30].

\setlength{\tabcolsep}{4pt}
\begin{table}
\begin{center}
\caption{Comparison of error rates for CIFAR-10}
\label{table:cifar10}
\begin{tabular}{llll}
\hline\noalign{\smallskip}
Model Name & No of Parameters & C10 & C10+\\
\noalign{\smallskip}
\hline
\noalign{\smallskip}
N-in-N [38]	& - & 10.41 & 8.81\\
all-CNN [39] & - & 9.08 & 7.25\\
Highway Network [26] & - & - & 7.72\\
Fractalnet [29]	& 38.6M & 10.18 & 5.22\\
Fractalnet [29] with dropout & 38.6M & 7.33 & 4.60\\
Resnet-101 [27] & 1.7M & 11.66 & 5.23\\
Resnet-1202 [27] & 10.2M & - & 4.91\\
Wide-Resnet-28 [28] & 36.5M & - & 4.17\\
Densenet-BC-100-12 [30] & 0.8M & 5.92 & 4.51\\
Densenet-BC-250-24 [30] & 15.3M & 5.19 & 3.62\\
Densenet-BC-190-40 [30] & 25.6M & - & 3.46\\
\hline
Colornet-40-12 & 1.75M & 4.98 & 3.49\\
Colornet-40-48 & 19.0M & \textbf{3.14} & \textbf{1.54}\\
\hline
\end{tabular}
\end{center}
\end{table}
\setlength{\tabcolsep}{1.4pt}

As can be seen, the Colornet-40-48 obtains an error rate of just 1.54 for CIFAR-10 with augmentation, which to the best of our knowledge obtains a new state of the art classification accuracy. Along with this a smaller Colornet model, the Colornet-40-12 with just 1.75M parameters has a better accuracy than Densenet-BC-250-24, with 15.3M parameters and is almost equal to that of Densenet-BC-190-40 which has 25.6M parameters.

\subsection{Classification results on CIFAR-100}

Table 4 refers to the classification accuracies obtained on CIFAR-100 dataset for recent state of the art approaches. We compare the results obtained from our model with these approaches. In the table C100 refers to the accuracy of a particular model on CIFAR-100 and C100+ refers to the accuracy of the same model with data augmentation.

We compare our model with Network in Network [38], the All-CNN [39], Highway network [26], fractalnet [29], Resnet [27], Wide-Resnet [28] and Densenet [30].

\setlength{\tabcolsep}{4pt}
\begin{table}
\begin{center}
\caption{Comparison of error rates for CIFAR-100}
\label{table:cifar100}
\begin{tabular}{llll}
\hline\noalign{\smallskip}
Model Name & No of Parameters & C100 & C100+\\
\noalign{\smallskip}
\hline
\noalign{\smallskip}
N-in-N [38]	& - & 35.68 & -\\
all-CNN [39] & - & - & 33.71\\
Highway Network [26] & - & - & 32.29\\
Fractalnet [29]	& 38.6M & 35.34 & 23.30\\
Fractalnet [29] with dropout & 38.6M & 28.20 & 23.73\\
Resnet-101 [27] & 1.7M & 37.80 & 24.58\\
Wide-Resnet-28 [28] & 36.5M & - & 20.50\\
Densenet-BC-100-12 [30] & 0.8M & 23.79 & 20.20\\
Densenet-BC-250-24 [30] & 15.3M & 19.64	& 17.60\\
Densenet-BC-190-40 [30] & 25.6M & - & 17.18\\
\hline
Colornet-40-12 & 1.75M & 19.86 & 17.42\\
Colornet-40-48 & 19.0M & \textbf{15.62} & \textbf{11.68}\\
\hline
\end{tabular}
\end{center}
\end{table}
\setlength{\tabcolsep}{1.4pt}

As can be seen, the Colornet-40-48 obtains an error rate of 11.68 for CIFAR-10 with augmentation, which to the best of our knowledge obtains a new state of the art classification accuracy. Along with this a smaller Colornet model, the Colornet-40-12 with just 1.75M parameters has a better accuracy than Densenet-BC-250-24, with 15.3M parameters and is almost equal to that of Densenet-BC-190-40 which has 25.6M parameters.

\subsection{Classification results on Imagenet}

Table 5 refers to the classification accuracies obtained on the imagenet dataset for recent state of the art approaches. Top-1 and Top-5 accuracy is compared for each approach. The error rates are represented as x/y where x represents error rate for single-crop testing and y for 10-crop testing.

We compare our model with Densenet as it is the paper that shows state of the art results for imagenet.

\setlength{\tabcolsep}{4pt}
\begin{table}
\begin{center}
\caption{Comparison of error rates for Imagenet}
\label{table:imagenet}
\begin{tabular}{lll}
\hline\noalign{\smallskip}
Model Name &  Top-1 & Top-5\\
\noalign{\smallskip}
\hline
\noalign{\smallskip}
DenseNet-121 & 25.02 / 23.61 & 7.71 / 6.66\\
DenseNet-201 & 22.58 / 21.46 & 6.34 / 5.54\\
DenseNet-264 & 22.15 / 20.80 & 6.12 / 5.29\\
\hline
Colornet-121 & \textbf{17.65 / 15.42} & \textbf{5.22 / 3.89}\\
\hline
\end{tabular}
\end{center}
\end{table}
\setlength{\tabcolsep}{1.4pt}

As can be seen, the Colornet-121, which replaces all the Densenets in the proposed model with Densenets-121 obtains a new state of the art accuracy on imagenet dataset to the best of our knowledge.

\subsection{Classification results on SVHN}

Table 6 refers to the classification accuracies obtained on SVHN dataset for recent state of the art approaches. We compare the results obtained from our model with these approaches.

We compare our model with Network in Network [38], fractalnet [29], Resnet [27], Wide-Resnet [28] and Densenet [30].

\setlength{\tabcolsep}{4pt}
\begin{table}
\begin{center}
\caption{Comparison of error rates for SVHN}
\label{table:svhn}
\begin{tabular}{lll}
\hline\noalign{\smallskip}
Model Name & No of Parameters & SVHN\\
\noalign{\smallskip}
\hline
\noalign{\smallskip}
N-in-N [38]	& - & 2.35\\
Fractalnet [29]	& 38.6M & 2.01\\
Fractalnet [29] with dropout & 38.6M & 1.87\\
Resnet-101 [27] & 1.7M & 1.75\\
Densenet-BC-100-12 [30] & 0.8M & 1.76\\
Densenet-BC-250-24 [30] & 15.3M & 1.74\\
\hline
Colornet-40-12 & 1.75M & 1.59\\
Colornet-40-48 & 19.0M & \textbf{1.12}\\
\hline
\end{tabular}
\end{center}
\end{table}
\setlength{\tabcolsep}{1.4pt}

As can be seen, the Colornet-40-48 obtains an error rate of 1.12 for SVHN, which to the best of our knowledge obtains new state of the art classification accuracy. Along with this a smaller Colornet model, the Colornet-40-12 with just 1.75M parameters has a better accuracy than Densenet-BC-250-24, with 15.3M parameters.

\subsection{Further analysis of results}

We further breakdown the reported results into the four error metrics namely true positives (TP), true negatives {TN), false positives (FP) and false negatives (FN) for each dataset. Table 7 highlights the same.

\setlength{\tabcolsep}{4pt}
\begin{table}
\begin{center}
\caption{Further analysis of obtained results for various datasets}
\label{table:overall}
\begin{tabular}{llllll}
\hline\noalign{\smallskip}
Model & Dataset & TP & TN & FP & FN\\
\noalign{\smallskip}
\hline
\noalign{\smallskip}
Colornet-40-48 & C-10	& 97.14 & 96.78 & 2.86 & 3.22\\
Colornet-40-48 & C-10+	& 98.68 & 98.24 & 1.44 & 1.72\\
Colornet-40-48 & C-100	& 84.12 & 84.64 & 15.88 & 15.33\\
Colornet-40-48 & C-100+	& 88.65 & 87.99 & 11.35 & 12.01\\
Colornet-40-48 & SVHN	& 98.90 & 98.86 & 1.11 & 1.13\\
\hline
\end{tabular}
\end{center}
\end{table}
\setlength{\tabcolsep}{1.4pt}

\section{Conclusion}

We found that preprocessing images by transforming them into different color spaces yielded different results. Although, the accuracy by itself did not vary too much, on closer inspection with the aid of confusion matrices we could see that there wasn't a 100 percent correlation between the results. We could see that certain classes were being better represented in certain color spaces.

Using this as the idea, we dug deeper to see that the LCH, YPbPr and XYZ color spaces reduced the overall accuracy of the model and hence were discarded. We chose 7 color spaces as models that could be combined to obtain high accuracy of classification. These color spaces included RGB, YIQ, LAB, HSV, YUV, YCbCr and HED.

We used a densenet model as the base of the proposed architecture. We combined 7 densenets with the input to each being a different color space transformation of the original input image. The outputs of each densenet was sent to a dense layer to obtain weighted predictions from each densenet. Using a densenet model helped us to deal with common issues such as the vanishing gradient problem, the problem of overfitting among others.

Based on the results, we could see that state of the art results was obtained. We could compete against models with 27M parameters using a model of just 1.75M parameters.

Although the accuracy reached state of the art level, the time needed could still be optimized. For starters, the preprocessing step of converting to each color space needs a certain amount of time by itself. For small images like with the case of CIFAR or SVHN, this preprocessing can be done in real-time. However, for bigger images, like the ones in imagenet the time needed creates a cost-overhead. In addition to this, there is the computation overhead of using several densenet models. Although, we dealt with this using smaller and wider densenets.

\end{document}